\documentclass[11pt,a4paper]{article}
\usepackage[breaklinks=true]{hyperref}
\usepackage{acl2021}
\usepackage{times}
\usepackage{latexsym}
\usepackage{amssymb}

\usepackage{microtype}

\aclfinalcopy 

\title{Modeling Users and Online Communities for Abuse Detection: \\A Position on Ethics and Explainability}

\author{Pushkar Mishra$^\bigstar$,\,\, Helen Yannakoudakis$^\spadesuit$,\,\, Ekaterina Shutova$^\clubsuit$\\
  $^\bigstar$ Facebook AI, London, United Kingdom\\
  $^\spadesuit$ Department of Informatics, King's College London, United Kingdom\\
  $^\clubsuit$ Institute for Logic, Language and Computation, University of Amsterdam, The Netherlands\\
  \normalsize{{\tt pushkarmishra@fb.com, helen.yannakoudakis@kcl.ac.uk, e.shutova@uva.nl}}
}

\date{}

\begin{document}
\maketitle
\begin{abstract}
Abuse on the Internet is an important societal problem of our time. Millions of Internet users face harassment, racism, personal attacks, and other types of abuse across various platforms. The psychological effects of abuse on individuals can be profound and lasting. Consequently, over the past few years, there has been a substantial research effort towards automated abusive language detection in the field of NLP. In this position paper, we discuss the role that modeling of users and online communities plays in abuse detection. Specifically, we review and analyze the state of the art methods that leverage user or community information to enhance the understanding and detection of abusive language. We then explore the ethical challenges of incorporating user and community information, laying out considerations to guide future research. Finally, we address the topic of explainability in abusive language detection, proposing properties that an explainable method should aim to exhibit. We describe how user and community information can facilitate the realization of these properties and discuss the effective operationalization of explainability in view of the properties.
\end{abstract}

\section{Introduction}
\label{intro}
With the advent of social media, anti-social and abusive behavior has become a prominent occurrence online. Undesirable psychological effects of abuse on individuals make it an important societal problem of our time. Munro \shortcite{munro2011} studied the ill-effects of online abuse on children, concluding that children may develop depression, anxiety, and other mental health problems as a result of their encounters online. \textit{Pew Research Center}, in its latest report on online harassment \cite{pew}, revealed that $40\%$ of adults in the United States have experienced abusive behavior online, of which $18\%$ have faced severe forms of harassment, e.g., that of sexual nature. These statistics stress the need for automated detection and moderation systems. Hence, in recent years, a new research effort on abusive language detection has sprung up in NLP.

That said, the notion of abuse has proven elusive and difficult to formalize. Different norms across different (online) platforms can affect what is considered abusive \cite{chandrasekharan2018internet}. In the context of natural language, \textit{abuse} is a term that encompasses many different fine-grained types of negative expressions. For example, Nobata et al. \shortcite{nobata} use it to collectively refer to hate speech, derogatory language and profanity, while Mishra et al. \shortcite{mishra} use it to discuss racism and sexism. The definitions for different types of abuse tend to be overlapping and ambiguous. However, regardless of the specific type, we define abuse as \textit{any expression that is meant to denigrate or offend a particular person or group}.
Taking a course-grained view, Waseem et al. \shortcite{W17-3012} classify abuse into broad categories based on \textit{explicitness} and \textit{directness}. \textit{Explicit} abuse comes in the form of expletives, derogatory words or threats, while \textit{implicit} abuse has a more subtle appearance characterized by the presence of ambiguous terms and figures of speech such as metaphor or sarcasm. \textit{Directed} abuse targets a particular individual as opposed to \textit{generalized} abuse which is aimed at a larger group such as a particular gender or ethnicity.

To date, several approaches to automated detection of abusive language have been proposed, including rule-based \cite{smokey,razavi,wiegand}, linguistic and social feature engineering \cite{Yin09detectionof,sood2012,warner,salminen2018anatomy}, utilizing distributed representations from neural networks \cite{djuric,mehdad,nobata} or applying deep neural networks directly \cite{park_fung,pavlopoulos-emnlp,mishra}. Researchers have also explored multi-task learning settings with objectives such as emotion detection \cite{santosh,samghabadi2019attending}. We refer the reader to recent surveys of the field \cite{schmidt_wiegand,fortuna_survey} for a detailed literature review.

More recently, researchers have noted that the linguistic features of a comment alone may not be sufficient to classify it as abusive or not. Information of the user who posted the comment, and of the surrounding social community of that user, further provides valuable insights into the abusiveness of the comment. An example of this is the study by Zook \shortcite{zook}, which mapped the locations of racist tweets in response to President Obama's re-election to show that such tweets were not uniformly distributed across the United States but instead came from specific geographical communities of users. Other works have also shown how users on online platforms organize into communities based on factors such as shared beliefs, stereotypes, linguistic norms, or geographical propinquity \cite{ICWSM136067,10.5555/2021109.2021119}.

In this paper, we focus on the role that modeling of users and communities plays in the automated detection of abusive language on online platforms. Specifically, we investigate the different state of the art methods that leverage user or community information to enhance the understanding and detection of abusive language. While these methods have yielded high performance gains, there has been little discussion of the kinds of information they capture. We provide a comprehensive review of these methods, analyzing the information they encode about users or communities and the relevance of that for detection of abusive language. We then explore the ethical considerations of incorporating user and community information in such methods, providing guidance for future research. Finally, we address the topic of explainability in abusive language detection, proposing properties that an explainable detection method should aim to exhibit. We describe how user and community information can facilitate the realization of these properties and discuss the effective operationalization of explainability in view of the properties.


\section{Why the user and community matter}
\label{why-user-community-matters}
Throughout the paper, \textit{user} refers to the user of an online platform who may have posted a comment that is to be classified as abusive or not. The \textit{community} of this user comprises other users and contents that they interact with on the online platform. In other words, community refers to the neighborhood of the user in the social graph of the platform. Conversations online are inherently contextual. Consequently, abuse on online platforms can only be effectively interpreted within a larger \textit{context} \cite{gao-fox} rather than in isolation. This is especially true for implicit or generalized abuse, which are harder to interpret than explicit abuse for humans and machines alike. Information of the user who posted the comment, or of the surrounding community including the targets of the comment, offers insights into several aspects of the context that are otherwise not accessible through the linguistic content of the comment alone. Here, \textit{information} may refer to demographic traits like age or gender, knowledge about linguistic behavior, location details, etc. Below we categorize and discuss the aspects of the context relevant to abusive language detection.

\vspace{1mm}
\noindent
\textbf{Sociolinguistic norms.} \textit{Sociolinguistics} studies the effects of society on language and its usage. Researchers in the past have explored the links between the structures and norms of real-world communities and the linguistic practices of people \cite{socio}. As in the physical world, individuals and communities on online platforms also abide by certain norms, which may be guided by their cultural backgrounds and/or are based on the standards laid down by the platforms themselves. These norms and standards reflect expectations of \textit{respectful} behavior, local customs and language patterns within a region, etc. \cite{IJoC3697}. Consequently, the decision of what is considered abusive must be made taking into account the sociolinguistic norms. User and community information, when leveraged alongside linguistic features, helps capture the relevant sociolinguistic norms in a myriad of ways. For example, a comment may contain the \textit{n}-word, but interpretation of its use and or the intent is greatly facilitated by the knowledge of the ethnicity of the user who wrote the comment and/or the ethnicity of the target user or community.

\vspace{1mm}
\noindent
\textbf{Linguistic variations.} Another aspect comes from looking at implicit abuse, whereby a user may utilize novel \textit{slangs} or conventional words in unconventional ways, e.g., as a racial slur or as a name for some specific demographic \cite{W17-3012}. Information about how a term is being used by other members of a user's community, e.g., in abusive contexts or otherwise, can help decipher linguistic variations that come up from time to time. In fact, it is usually the users with strong ties who are responsible for popularizing language variations as well as for spreading hate speech \cite{del-tredici-fernandez-2018-road,ribeiro}. Therefore, having user and community information alongside linguistic features helps capture linguistic variations and their diffusion.

\vspace{1mm}
\noindent
\textbf{Prevailing stereotypes.} Previous research has shown that prevailing stereotypes often form the basis and justification of abuse. For example, many twitter accounts were open about their anger and hatred for Muslims in the wake of the Rochdale scandal that involved several British--Asian men getting convicted for child grooming \cite{Awan2014IslamophobiaAT}. Stereotypes are not only explicit but implicit too \cite{Hinton2017ImplicitSA}, which often show up as implicit and subtle abuse in the form of sarcasm, racist jokes, or unnecessary associations. While explicit stereotypes are consciously endorsed, and may be controllable, implicit stereotypes are thought to be shaped by experience and based on learned associations \cite{Byrd2019WhatWC}. User and community information plays an important role in the identification of such stereotypes. For example, if the location of users is available alongside linguistic features of the comments they post, one can quickly discover the presence (or absence) of correlations between specific regions and specific kinds of abuse. Moreover, shared stereotypes may unconsciously bring users together on online platforms to form communities. Hence, having linguistic information of a community, such as the topics users in that community interact with and the stance of users towards different pieces of news, can help capture the prevailing stereotypes that form the motivation behind abusive comments from such users.

\vspace{1mm}
\noindent
\textbf{Demographic characteristics.} Previous research has demonstrated that some demographic settings are inherently more abusive than others. For example, a study by Stephens et al. \shortcite{stephens} mapped the locations of hateful tweets across the United States to uncover the regions where people use hate speech the most. They observed that areas with low diversity use more derogatory slurs against racial and sexual minorities. A separate line of work by Savicki et al. \shortcite{savicki} concluded that male-only discussion groups on the Internet use more coarse and abusive language than female-only groups. These works indicate that demographic settings can be predictive of the (abusive) nature of comments originating from within them. User and community information constitutes a direct and simple way of capturing the demographic setting of a comment.


\section{Modeling the user and community}
Existing methods for abusive language detection that leverage information of the user who posted the comment or their community can be categorized as social feature engineering based, user embedding based, and social graph based approaches.

\vspace{1mm}
\noindent
\textbf{Social feature engineering based.} These methods directly incorporate hand-engineered features and personal traits of users or their communities in order to model the likelihood of abusive language in the users' comments, a process known as \textit{profiling} \cite{zhang2018natural}. Dadvar et al. \shortcite{davdar} included the age of users alongside other traditional lexicon-based features to detect cyber-bullying, while Galán-García et al. \shortcite{galan2016supervised} utilized the time of publication, geo-position and language in the profile of Twitter users. Waseem and Hovy \shortcite{waseem_hovy} exploited gender of Twitter users on top of character n-gram counts to improve detection of sexism and racism in a dataset comprising racist, sexist and benign tweets -- they noted that the F$_1$ increased slightly from $73.89\%$ to $73.93\%$ when the gender feature was included. Using the same setup, Unsvåg and Gambäck \shortcite{unsvaag2018effects} showed that the inclusion of social community (i.e., number of followers and friends) and activity (i.e., number of status updates and favorites) features of users alongside their gender further enhanced performance by $3$ F$_1$ points over the n-gram baseline.

\vspace{1mm}
\noindent
\textbf{User embeddings based.} These methods utilize neural networks to generate representations, called \textit{profiles}, for users that capture their linguistic behavior based on the comments they created. Pavlopoulos et al. \shortcite{W17-4209} worked with a dataset of abusive and benign comments in Greek provided by the news portal \textit{Gazzetta}. They divided the users whose comments are in the dataset into four \textit{types} based on the proportion of abusive comments: \textit{red} users (e.g., if $>10$ comments and $\geq 66\%$ abusive comments), \textit{yellow} users (with $>10$ comments and $33\%-66\%$ abusive comments), \textit{green} users (with $>10$ comments and $\leq 33\%$ abusive comments), and \textit{unknown} users (users with $\leq 10$ comments). They then assigned unique randomly-initialized embeddings to users and added them as additional input alongside representations of comments obtained from the GRU model of Pavlopoulos et al. \shortcite{pavlopoulos-emnlp}. This increased the AUROC from $79.24\%$ to $80.71\%$. Qian et al. \shortcite{N18-2019} used LSTMs to model the inter and intra-user relationships in the dataset by Waseem and Hovy \shortcite{waseem_hovy} with sexist and racist tweets combined into one category. They first applied a bi-LSTM to users' recent tweets in order to generate intra-user representations that capture the history of their content. To improve robustness against the noise present in tweets, they then utilized locality sensitive hashing to form sets of semantically similar tweets. They trained a policy network to select tweets from these sets that a bi-LSTM could use to generate inter-user representations. When these inter and intra-user representations were utilized alongside representations of tweets from a bi-LSTM baseline, the F$_1$ score increased from $70.3\%$ to $77.4\%$.

\vspace{1mm}
\noindent
\textbf{Social graph based.} These methods leverage the social relations (e.g., friendship) that exist amongst users in a social network. Mishra et al. \shortcite{mishra} constructed a social graph of all the users whose tweets are in the dataset of Waseem and Hovy \cite{waseem_hovy}. Nodes were the users and edges the follower--following relationship amongst them on \textit{Twitter}. The researchers applied \textit{node2vec} \cite{node2vec-kdd2016} to this graph to generate representations for users, i.e., \textit{profiles}, which capture information about their social connections. The addition of these profiles on top of linguistic representations of tweets yielded significant gains whereby the F$_1$ scores on the racism and sexism classes increased from $72.28\%$ and $72.09\%$ to $75.09\%$ and $82.75\%$ respectively. The gains were attributed to the fact that the profiles captured not only information about respective communities of users but also enabled modeling of the topical contexts amongst the connected users. Mishra et al. \shortcite{mishragcn} further expanded on this work by adding tweet nodes to the social graph of Mishra et al. \shortcite{mishra} alongside user nodes. They connected every tweet node to the corresponding user who posted the tweet. They then used a graph convolutional network \cite{gcn} to create profiles of users that now captured their linguistic behavior too. When they used these profiles together with the linguistic representations of tweets, F$_1$ scores on the racism and sexism classes further improved to $79.49\%$ and $84.44\%$ respectively. Ribeiro et al. \shortcite{ribeiro} also applied graph neural networks, GraphSage \cite{graphsage}, to a social graph of ca. $100k$ Twitter users to generate profiles that they used to classify the users as hateful or normal. They noted that their social graph based method outperformed traditional gradient-boosted decision tree classifiers by $15$ F$_1$ points on the same task. Tredici et al. \shortcite{del-tredici-etal-2019-shall} constructed a graph of users whose tweets are in the hate-speech dataset of Founta et al. \shortcite{ICWSM1817909}. Nodes were uses and edges between them signified that one user retweeted the other. They used Graph Attention Networks \cite{gat} to generate representations of users from this graph, which when used alongside linguistic representations, provided a gain of 5 $F_1$ points. Cecillon et al. \shortcite{Cecillon2021GraphEF} worked with a social graph of users from a French gaming website where weighted edges represented the intensity of communication between the users. Then for each comment to be classified, the researchers extracted the ego-graph of its author and created a feature vector for the comment from the ego-graph using \textit{node2vec} along with measures like degree centrality. An SVM trained with these graph-based feature vectors reached 89 F$_1$ points as opposed to 81 F$_1$ points when trained with content features.


\section{Analysis of the methods}
\label{methods}
We now analyze the methods described above to understand the gains that user or community information provides. Based on this analysis, in the next sections, we explore the ethical considerations of incorporating user and community information and how it can support explainability.

Across the three categories of methods, we note that the general setup is to create representations, called \textit{profiles}, for users or communities and utilize them alongside linguistic features. In social feature engineering based methods, these profiles are manually constructed vectors of features that capture the relevant traits, such as age in the case of cyber-bullying and gender in the case of sexism. In user embeddings and social graph based methods, the profiles are instead generated by neural network architectures to capture the linguistic behavior or community traits of users. That said, across all three categories, the profiles essentially provide a wider context to the comment being classified for abuse. For example, having the gender of the user who produces a comment such as ``\textit{Had an accident, women can't drive it seems!}" can help to classify the comment as sexist or not by differentiating benign self-deprecating humor from intent to degrade. The context that the profiles encode increases as we go from social feature engineering based methods to user embeddings based methods and further to social graph based methods. This is also evident from the magnitude of gains that the profiles provide on top of linguistic features. For example, the gender feature only increases the F$_1$ from $73.89\%$ to $73.93\%$ over character n-gram counts on the dataset by Waseem and Hovy \shortcite{waseem_hovy}, while the social graph based method of Mishra et al. \shortcite{mishragcn} increases the F$_1$ to above $80\%$. The example aside, it makes intuitive sense that profiles from social graph based methods encode the most amount of context, since these profiles are able to capture the various phenomena that occur in social networks, the most prominent ones of which are:
\begin{itemize}
    \item \textit{Homophily}, i.e., the tendency of users in a social space forge ties with others who are similar to them in socially significant ways \cite{homophily}.
    \item \textit{Coordinated behavior} or \textit{brigading}, i.e., when users with similar beliefs act in a coordinated manner in a social space towards some common objective \cite{Parent2019SocialMB}.
\end{itemize}

\noindent
In fact, homophily is so prominent, Mishra et al. \shortcite{mishragcn} noted in their work that the profiles they generated from the social graph of users and tweets could encode patters of similar linguistic practices amongst connected users in the Waseem and Hovy \shortcite{waseem_hovy} dataset, hence allowing for comments with implicit and generalized sexism or racism to be better detected. Moreover, homophily has direct associations with all the four aspects of context that we described in section \ref{why-user-community-matters}, i.e., similar sociolinguistic norms and shared language markers facilitate homophilic ties in social networks \cite{doi:10.1177/0956797619894557}, as do shared beliefs, stereotypes, and demographic traits \cite{mishra}. Therefore, capturing homophily allows for all the four aspects to be directly captured together.

We note that just exploiting simplistic and limited inductive biases that are easy to extract, like gender of the user, can render methods prone to making faulty generalizations because of over-fitting to patterns in the training data. This is also evident from the observations that Mishra et al. \shortcite{mishragcn} made in their work. They noted that the profiles they generated from the social graph consisting of user and tweet nodes improved F$_1$ scores over the profiles Mishra et al. \shortcite{mishra} generated from the social graph just consisting of users, with the gains mainly coming from increase in precision. It is because solely relying on \textit{network homophily} as the inductive bias for generating profiles caused the method of Mishra et al. \shortcite{mishra} to make some faulty generalizations. Such observations have also been made by other works, a prominent one of which is the work of Bamman et al. \shortcite{Bamman} who explored the relationships amongst gender, language, and social network connections. The researchers noted that even though there may exist many linguistic clusters that exhibit strong orientations to one gender, yet the characteristics of any particular cluster do not necessarily align with population-level statistics for that gender. Furthermore, they observed that there are individuals whose linguistic practices differ from population-level trends for their gender and that gender homophily does not capture their linguistic practices.

\section{Ethical considerations}
While researchers have started incorporating user and community information into detection of abusive language, there has been no discussion of the ethical guidelines for doing so. Therefore, taking a stand on the issue, we lay out five ethical considerations in the design and implementation of methods that incorporate user or community information:

\vspace{1mm}
\noindent
\textbf{Personal vs. population-level trends.} It is important to perform appropriate generalizations from personal traits to population-level behavioral trends. Methods should avoid relying on simple inductive biases such as personal traits of users, e.g., gender, race, etc., as this can easily lead to scenarios of faulty generalizations where comments from a particular gender or race are always labeled abusive/benign. Moreover, relying solely on personal traits of users also comes with the risk that such information may not always be present or may not be accurate even when present \cite{10.1016/j.chb.2016.06.052}. On the other hand, more complex inductive biases learned from data, as in the case of social graph based methods, provide a safer and more reliable generalization from personal behaviors of users or communities to population level trends.

\vspace{1mm}
\noindent
\textbf{Bias in datasets.} An obvious pitfall in working with methods that incorporate user and community information is having datasets where comments come from users belonging to some limited demographics only. We refer to this as \textit{demographic bias}. Datasets with demographic bias will cause the methods to overfit to linguistic practices and dialects of users and communities belonging to specific demographics \cite{sap-etal-2020-social}, hence diminishing the power of the methods to generalize. In fact, this bias is not only a problem for methods we discussed, but for any NLP method in general. When it comes to methods that incorporate user or community information specifically, there are two other biases that must be kept in mind when constructing datasets; we refer to them as \textit{comment distribution bias} and \textit{label distribution bias}. Comment distribution bias occurs when the majority of comments in the dataset come from a small number of unique users. Such datasets allow the methods to simply overfit to the linguistic or social behaviors and community roles of specific users \cite{wiegand-etal-2019-detection}. Label distribution bias occurs when only the abusive comments of a user are included in the dataset. Abuse is a relatively infrequent phenomenon, even at an individual level \cite{waseem_hovy, wulczyn}. Only getting abusive comments of a user can make the methods simply associate the identity of the user to abusiveness when including user information. Moreover, datasets with this bias can also make phenomena like homophily appear overly effective in the detection of abuse by sampling only abusive comments from users who are close in the social network.

\vspace{1mm}
\noindent
\textbf{Observability.} The observability aspect needs to be accounted for, i.e., does a method allow for the profiling knowledge it has learned about users and communities to be directly or indirectly observed in its workings, e.g., if it has segregated users into categories observable by others. If yes, that can be used as a basis for systematic oppression of certain users or communities by other users and communities. A prime example of this is when users report benign comments that they do not agree with as abusive since they have noted that the detection method is more likely to adjudicate the comments abusive simply because they come from a particular community or a particular user.

\vspace{1mm}
\noindent
\textbf{Privacy.} As we discussed in the previous section, profiles created by the methods may carry a lot of information about the personal traits of users, their linguistic practices, etc. Furthermore, the information carried increases in specificity as we go from social feature engineering based methods to social graph based methods. An important ethical consideration that then arises is whether the profiles or the models learned by the methods be made available publicly. Doing so may allow for users and communities to be uniquely identified and for their sociolinguistic behaviors, community roles, or personal and population-level beliefs to be exposed.

\vspace{1mm}
\noindent
\textbf{Purpose.} The purpose of leveraging user and community information should be made clear upfront. Methods that leverage user and community information to enhance the detection of abusive language in comments should be preferred over those that leverage the information to classify users or communities themselves as abusive. This is because the latter can lead to unwarranted penalties, e.g., a platform may prohibit a user from engaging even in restorative conversations simply because of their past abusive behavior.

\section{Explainable abusive language detection}
\textit{Explainability} is an important concept within abusive language detection. Jurgens et al. \shortcite{jurgens} noted in their work that explainable ML techniques can promote \textit{restorative} and \textit{procedural} justice by surfacing the norms that have been violated and clarifying how they have been violated. That said, there has been limited discussion of the issue within the domain of abusive language detection. In this section, we first formalize the properties that an explainable detection method should aim to exhibit in order to thoroughly substantiate its decisions. We then describe how user and community information play an important role in the realization of each of the properties. Finally, we discuss what it means to operationalize explainability within abusive language detection in an effective manner.

\subsection{Properties of an explainable method}
In drawing up the properties that an explainable method for abusive language detection should aim to exhibit, we take into account the taxonomy of abuse we discussed in the introduction, i.e., directed vs. generalized and implicit vs. explicit:

\begin{itemize}
    \item \textit{Provide evidence for intent} of abuse (or the lack of it), hence convincingly segregating abuse from other phenomena such as sarcasm and humor.
    \item \textit{Point out the abusive phrases} within a comment (or the absence thereof), be they explicit (e.g., expletives or slurs) or implicit (e.g., dehumanizing comparisons).
    \item \textit{Identify the target(s)} of abuse (or the absence thereof), be it an individual (i.e., directed abuse) or a group (i.e., generalized abuse).
    \item \textit{Elucidate stereotypes(s)} underlying the abuse (or the absence thereof), be they explicit or be they in the form of implicit associations.
\end{itemize}

\noindent
User and community information has a crucial role to play in the effective realization of each of the four properties. For the first property, as illustrated earlier in the paper, information of the user who created the comment can serve as evidence for whether the comment intends to be degrading to others or just self-deprecating humor. For the second property, let us consider a comment like ``\textit{You're a \underline{pig}!}"; if directed at people belonging to certain religions, it may constitute an implicit racial slur, but otherwise, may simply be viewed as a remark on cleanliness. So, the information of the user or community being targeted can explain whether a phrase is abusive or not. The methods we analyzed in section \ref{methods} do not model the information of the target user or community, which is a valuable direction for future research. For the third property, we note that social graph based methods are inherently suited to provide a convenient setup for identification of the user or community being targeted by an abusive comment, specially in scenarios where the social graph is enriched with information like the topics being discussed amongst groups of connected users. For the fourth property, user and community information again offers a direct way to elucidate explicit or implicit stereotypes, e.g., by exposing the associations being made by a community between certain qualities and the targets of their abuse.

\subsection{Operationalizing explainability}
Having formalized the properties that an explainable detection method should aim to exhibit, we now address the question of how explainability can be effectively be operationalized within abusive language detection in view of these properties. We approach this discussion from three different perspectives, that of the designers of the detection method, that of the user creating comments, and that of the larger communities. By breaking the discussion down in this manner, we explore the different choices that exist for operationalization and the purposes they can serve.

\vspace{1mm}
\noindent
\textbf{Designers of the method.} For the designers of the detection method, explainability can serve as a principled mechanism for understanding and reasoning about the behavior of their method, which is important for multiple reasons. Firstly, if the detection method exhibits all the four properties of explainability, then the designers can easily gain insights into the factors that contributed to the decision made by the method given a comment. This can allow the designers to recognize when the method may be overly relying on a specific factor, e.g., the demographic traits. In the case of social feature engineering and user embeddings based methods, operationalization of explainability via feature attribution such as \textit{LIME} \cite{ribeiro-etal-2016-trust} and \textit{Integrated Gradients} \cite{sundararajan2017axiomatic} can be effective in offering such insights. For social graph based methods that employ graph neural networks, attribution techniques like \textit{GNNExplainer} \cite{gnnexplainer} can be used instead. The second reason why explainability is important for the designers is because it can allow them to optimize the method by removing inputs that do not contribute significantly. Here again, explainability via feature attribution can be effective. Lastly, explainability is also important for the designers to understand how their method would perform in cases where a user may try obfuscate abusive language \cite{nobata}. Counterfactual explanations can constitute an effective operationalization for the designers to identify the parts of their method that are most vulnerable to obfuscations.

\vspace{1mm}
\noindent
\textbf{Users.} Besides being a mechanism for designers to interpret their methods, an effective operationalization of explainability should also serve as a means for users to receive explanations for the decisions made by a detection method. Jurgens et al. \shortcite{jurgens} argue in their work that an online platform can build legitimacy and transparency by offering justifications to users when their comments are deemed abusive by the detection method of the platform, which can in turn lead to increase in compliance with the norms of the platform. That said, unlike in the case of designers of the method, offering feature attribution based explanations that simply highlight parts of a user's comment may not be effective at making the user agree with the decision of the detection method \cite{Carton_Mei_Resnick_2020}. Alternatively, providing a meaningful counterfactual paraphrase that is non-abusive is not only difficult \cite{laugel2019dangers}, but can also be seen as \textit{paternalism} on the part of the platform \cite{10.1145/3351095.3372830}, i.e., that the platform is trying to tell the user what to say or how to present their opinions. On the other hand, \textit{principal-reason explanations} \cite{10.1145/3351095.3372830}, whereby the detection method selects the reason(s) for its decision from a curated list, can constitute an effective operationalization. Such a list can be prepared for each of the four properties of explainability, e.g., by selecting the relevant norms from the \textit{terms of service} of the platform, hence allowing for a principal reason to be offered per property or a combination thereof. When coupled with feature attribution, this approach to operationalization can clearly indicate to the user the norm(s) that their comment violates and, where possible, highlight parts that contribute to the violation(s). For example, given a comment like ``\textit{You f***, why do you have to support that team??}", the detection method can highlight the first part based on feature attribution and select the norm forbidding the use of expletives directed at others.

\vspace{1mm}
\noindent
\textbf{Communities.} There can be scenarios where whole communities of users on a platform may be indulging in abusive behavior, e.g., by widely circulating an abusive view against a demographic group based on shared beliefs, common stereotypes or other homophilic ties. In such cases, just taking down specific instances of abusive language and providing justifications individually to the respective users may not prove effective. Users may continue to promote the abusive view, defying the norms of the platform in the process and ignoring the justifications given to them. The reason for this comes from \textit{social influence theory} which says that a user's behavior is affected by three broad varieties of social influence \cite{kelman}, i.e., \textit{compliance}, \textit{identification}, and \textit{internalization}. Compliance occurs when the user behaves a certain way so as to appear in congruence with opinions of others who matter to them; identification occurs when the user adopts behaviors in order to associate with others they admire; and internalization is when the user adopts the values and beliefs of others. The influences occur because of two needs of the user, the need to be liked (\textit{normative}) and the need to be right (\textit{informational}). In order to fulfill the latter, people may accept the three varieties of influence when there is lack of information, a concept known as \textit{social proof} \cite{cialdini2007influence}. Consequently, explainability has a bigger role to play here than simply being a tool that provides interpretability to designers or offers justifications to users. Operationalizing explainability in a manner that spreads awareness about existing stereotypes and fills the information gap can be very effective \cite{miller2018explanation,sap-etal-2020-social}. One way to achieve this is by having generative explanations in conjunction with information retrieval techniques that fulfill the property of elucidating stereotypes in a human-understandable way \cite{gilpin} while offering references to reliable sources on the stereotypes. In fact, such an operationalization that elucidates stereotypes or frames of bias \cite{sap-etal-2020-social} in abusive comments at a community level, while providing information to debunk the stereotypes themselves, can offer validation to the victims of abuse by communities, e.g., minority groups, and help them feel safer on the platform.


\section{Conclusions}
Abuse on the Internet stands as a significant challenge before the society. Its nature and characteristics constantly evolve, making it a complex phenomenon to study and model. In this paper, we explored the ways in which users and communities play a role in the detection of abusive language. We investigated the methods that leverage user or community information to uncover how they work and the knowledge they capture. We then explored the ethical challenges of incorporating user and community information, laying out considerations to guide future research. Finally, we moved to the topic of explainability in abusive language detection, proposing properties that an explainable detection method should aim to exhibit. We describe how user and community information can facilitate the realization of these properties and discussed the effective operationalization of explainability in view of the properties.

\bibliography{acl2021}
\bibliographystyle{acl_natbib}
\end{document}